\documentclass[conference]{IEEEtran}
\IEEEoverridecommandlockouts
\usepackage{cite}
\usepackage{xurl}      
\usepackage[hidelinks]{hyperref} 
\usepackage{amsmath,amssymb,amsfonts}
\usepackage{algorithmic}
\usepackage{graphicx}
\usepackage{textcomp}
\usepackage{array}
\usepackage{xcolor}
\def\BibTeX{{\rm B\kern-.05em{\sc i\kern-.025em b}\kern-.08em
    T\kern-.1667em\lower.7ex\hbox{E}\kern-.125emX}}
\begin{document}

\title{Automatic Landmark-Based Segmentation of Human Subcortical Structures in MRI\\

\thanks{$^*$Equal senior author contribution.}
}

\author{
Ahmed Rekik$^{1}$,
R. Jarrett Rushmore$^{2}$,
Sylvain Bouix$^{*1}$,
Linda Marrakchi-Kacem$^{*3}$\\[10pt]

$^{1}$École de technologie supérieure (ÉTS), Montréal, Canada\\
$^{2}$Boston University School of Medicine, Boston, MA, USA\\
$^{3}$Signal and Smart Systems Lab (L3S),\\ National School of Engineering of Tunis,\\ University of Tunis El Manar, Tunisia

}

\maketitle

\begin{abstract}
Precise segmentation of brain structures in magnetic resonance imaging (MRI) is essential for reliable neuroimaging analysis, yet voxel-wise deep models often yield anatomically inconsistent results that diverge from expert-defined boundaries. In this research, we propose a landmark-guided 3D brain segmentation approach that explicitly mimics the manual segmentation protocol of the Harvard--Oxford Atlas. A Global-to-Local network automatically detects 16 landmarks representing key subcortical reference points. Then, a semantic segmentation model produces a coarse segmentation of 12 anatomical labels, each grouping multiple subcortical regions. Finally, a landmark-driven post-processing step separates these 12 labels into 26 distinct structures by enforcing local anatomical constraints. Experimental results demonstrate consistent improvements in boundary accuracy. Overall, integrating learned landmarks aligns segmentations more closely with manual protocols.
\end{abstract}


\section{Introduction}
Accurate segmentation of subcortical structures in MRI is essential for neuroimaging research, clinical assessment, and neurosurgical planning \cite{Fischl2012}. Classical atlas-based methods, such as FSL-FIRST \cite{Patenaude2011} and multi-atlas label fusion approaches \cite{Iglesias2015}, incorporate anatomical priors through template registration and label propagation. Although these methods can provide anatomically consistent segmentations, they depend heavily on registration quality and may be limited in capturing subject-specific anatomical variability.
\\
Deep learning models such as U\textendash Net \cite{Ronneberger2015} and transformer\textendash based networks \cite{Chen2021,Yu2023} have significantly advanced automated segmentation. However, they may still produce anatomically inconsistent results -- for example, boundary leakage, fused neighboring structures, or deviations from expert neuroanatomical definitions \cite{Fu2021}.
\\
A major reason is that conventional neural networks rely primarily on voxel appearance (intensity, texture) without incorporating anatomical reference points \cite{hirsch2021segmentation}. In contrast, expert neuroanatomists rely on specific landmarks -- such as the anterior commissure (AC) or mammillary bodies -- to define where a structure begins, ends, or divides. These landmarks are essential for accurately and reliably segmenting complex regions that lie in close spatial proximity and \textit{often have no intensity difference} in T1-weighted (T1w) MRI.
\\
In the Harvard--Oxford Atlas (HOA) manual curation protocol~\cite{Rushmore2022,HOAManual2021}, each subcortical structure is delineated according to such landmarks and plane-based rules. This ensures consistent annotation across raters and datasets. However, manual annotation remains time-consuming and does not scale to large datasets. 
\\
In this paper, we translate this anatomical rigor into an automated framework by combining landmark detection with protocol-driven segmentation. 
This approach allows us to leverage the power of deep learning methods for automated semantic segmentation in areas where contrast and intensity patterns are informative, and also separate two or more structures using anatomical rules defined by external landmarks.

First, a Global-to-Local network detects 16 landmarks, following the HOA protocol. 
Second, a UNesT model~\cite{Yu2023} predicts a coarse segmentation with 12 merged labels. 
Finally, the coarse output is refined into 26 regions using landmark-defined planes that separate fused structures and
enforce neuroanatomical protocol constraints. Experiments show that the landmark-guided segmentation yields marginal global gains but achieves higher boundary accuracy and closer alignment with manual protocols.

\section{Methods}
\subsection{Codifying the HOA procedure}
\label{sec:hoacoding}

In accordance with the HOA protocols, 16 anatomical landmarks (Table~\ref{tab:landmarks_no_abbrev}) are defined to assist the manual delineation of 26 subcortical structures~\cite{HOAManual2021,Rushmore2022}.
For example, the anterior commissure (AC), posterior commissure (PC), and prepontine fissure (PPF) form a plane used to separate the left and right hemispheres. This novel approach bridges the gap between anatomical protocols defined by human experts and ``blind'', protocol-unaware semantic segmentation.
\begin{table}[htp]
\caption{Summary of the 16 anatomical landmarks}
\begin{center}
\begin{tabular}{|m{9em}|m{15em}|c|}
\hline
\textbf{Region} & \textbf{Landmark(s)} & \textbf{ID} \\
\hline
Putamen & First anterior appearance (L/R) & 1, 2 \\
\hline
Nucleus accumbens--putamen interface & Anterior contact (L/R) & 3, 4 \\
\hline
Nucleus accumbens--putamen interface & Posterior contact (L/R) & 5, 6 \\
\hline
Nubbins & Last anterior appearance of nucleus accumbens (L/R) & 7, 8 \\
\hline
Third ventricle & First anterior appearance & 9 \\
\hline
Commissures & Anterior / Posterior commissure & 10, 15 \\
\hline
Ventral Diencephalon & Mammillary bodies (L/R) & 11, 12 \\
\hline
Continuity of atrium & First posterior appearance of inferior horn (L/R) & 13, 14 \\
\hline
Brainstem limit & Prepontine fissure & 16 \\
\hline
\end{tabular}
\label{tab:landmarks_no_abbrev}
\end{center}
\end{table}

After reviewing this protocol, we observed that the process can be reproduced using a simplified formulation. Given 12 fused anatomical regions and 16 landmarks, the original 26-label ground truth can be reconstructed through deterministic post-processing rules (Table~\ref{tab:coarse_to_26}).

\begin{table*}[htbp]
\caption{Mapping from 26 to 12 fused anatomical labels}
\label{tab:coarse_to_26}
\begin{center}
\setlength{\tabcolsep}{4pt}
\renewcommand{\arraystretch}{1.15}
{
\begin{tabular}{|l|c|}
\hline
\textbf{Anatomical Structures (IDs; Abbreviations)} & \textbf{Fused Label} \\
\hline
(1, 2) Lateral Ventricle (LV\_L / LV\_R); (17, 18) Inferior Horn (IH\_L / IH\_R) & (1) LV + IH \\
\hline
(3) Cerebrospinal Fluid (CSF) & (2) CSF \\
\hline
(4) Third Ventricle (3V) & (3) 3V \\
\hline
(5) Fourth Ventricle (4V) & (4) 4V \\
\hline
(6, 7) Nucleus Accumbens (NAcc\_L / NAcc\_R); (10, 11) Putamen (Put\_L / Put\_R) & (5) NAcc + Put \\
\hline
(8, 9) Caudate (CAU\_L / CAU\_R) & (6) CAU \\
\hline
(12, 13) Globus Pallidus (GP\_L / GP\_R) & (7) GP \\
\hline
(14) Brainstem & (8) Brainstem \\
\hline
(15, 16) Thalamus (TH\_L / TH\_R) & (9) TH \\
\hline
(19, 20) Hippocampal Formation (HF\_L / HF\_R) & (10) HF \\
\hline
(21, 22) Amygdala (AMY\_L / AMY\_R) & (11) AMY \\
\hline
(23--26) Ventral Diencephalon (VDC\_A\_L/R, VDC\_P\_L/R) & (12) VDC \\
\hline
\end{tabular}}
\end{center}
\end{table*}

\begin{figure}[htbp]
\centering
\includegraphics[width=\columnwidth]{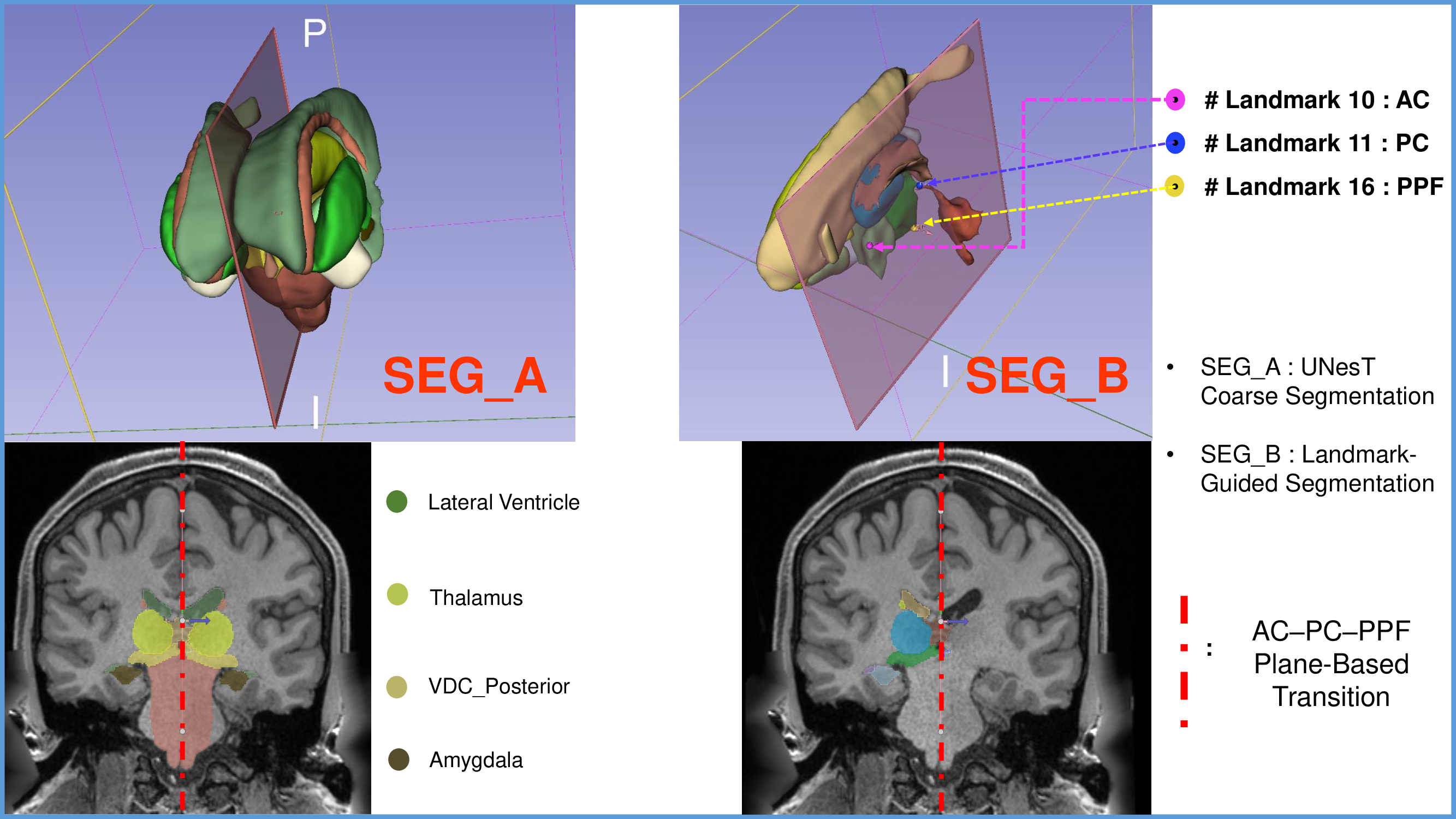}
\caption{Left/right separation using the midsagittal plane defined by AC, PC, and PPF. SEG\_A corresponds to the coarse segmentation produced by UNesT, while SEG\_B denotes the refined segmentation after landmark-guided post-processing using the AC–PC–PPF plane.}
\label{fig:acpc}

\end{figure}

\textbf{Midline splitting.}
A midsagittal plane defined by the AC, PC, and PPF splits each coronal slice into left and right hemispheres (Figure~\ref{fig:acpc}). For subsequent slices, the plane is slightly adjusted using the previous segmentation to correct interhemispheric misassignments.

\textbf{NAcc--Put separation.}
The fused NAcc--Put label is split using vertical slice-wise separators defined by the contact landmarks (\#3--\#6), yielding distinct NAcc and Putamen regions, as illustrated in Fig.~\ref{fig:expert_quals} (top).

\textbf{Coronal extents.}
Protocol-aligned boundaries are enforced using landmark-defined planes:
(i) the putamen is truncated anteriorly at landmarks \#1--\#2;
(ii) the NAcc is truncated posteriorly at landmarks \#7--\#8;
(iii) the third ventricle is excluded anterior to landmark \#9;
(iv) the VDC is divided into anterior and posterior parts by landmarks \#11--\#12;
and (v) the inferior horn is separated from the lateral ventricle up to landmarks \#13--\#14.
\subsection{Dataset}
We used the HOA subcortical dataset, consisting of 100 T1-weighted MRI volumes from the Human Connectome Project Young Adult cohort with manual voxel-wise segmentations of 26 subcortical brain structures~\cite{Rushmore2022}. The corresponding 16 anatomical landmarks were manually annotated in this study following the HOA protocol.

All scans have isotropic resolution of 0.7\,mm with matrix size $260 \times 311 \times 260$.

The dataset was split into 80 subjects for training, 10 for validation, and 10 for testing.

\textbf{Preprocessing.}
All MRI volumes were skull-stripped using SynthStrip~\cite{Hoopes2022} to remove non-brain tissue while preserving deep brain structures, followed by per-volume $z$-score intensity normalization, yielding consistent 3D inputs for both landmark detection and segmentation.

\subsection{Automated segmentation procedure}
\begin{figure}[htbp]
  \centering
  \includegraphics[width=\columnwidth]{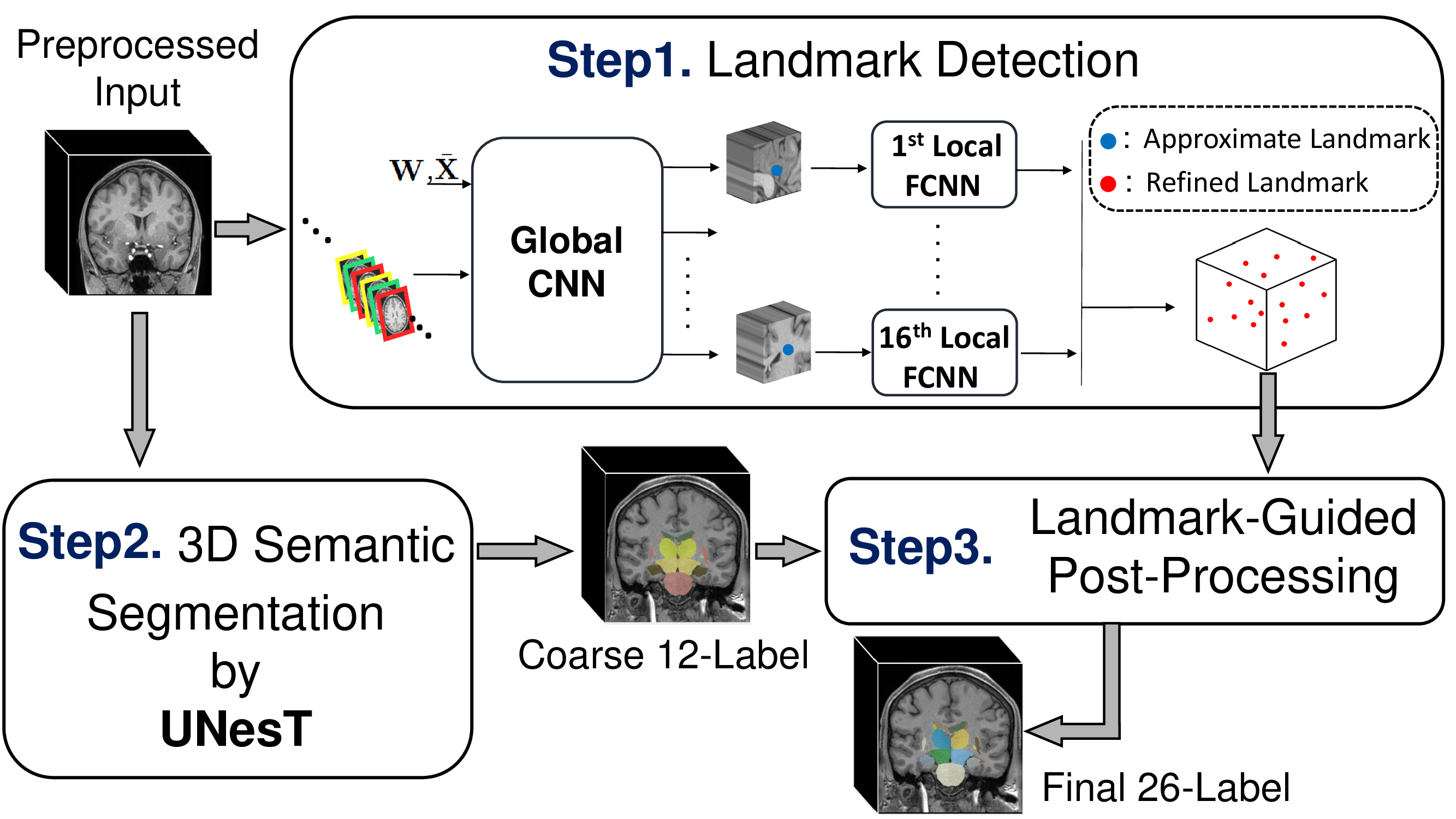}
  \caption{Overall pipeline.}
  \label{fig:pipeline}
\end{figure}

Our method, illustrated in Figure~\ref{fig:pipeline}, consists of three stages. We first detect 16 anatomical landmarks using a global-to-local network. Next, UNesT~\cite{Yu2023}, a 3D hierarchical transformer model, predicts a coarse 12-label segmentation. Finally, landmark-guided rules refine this output into 26 anatomically consistent regions (see Sec.~\ref{sec:hoacoding}).

\textbf{Step 1 -- Landmark detection.}
Accurate localization of neuroanatomical landmarks is critical for enforcing protocol-driven constraints. We use a global-to-local pipeline that embeds anatomical priors in a deep model.

\emph{Global model.}
We adopt the Patch-based Iterative Network (PIN)~\cite{Li2018PIN} to jointly estimate 16 landmarks. PIN localizes landmarks through an iterative process: at each iteration, a CNN predicts a displacement vector from the current patch center toward the target landmark, as well as a direction of movement. The patch is then recentered accordingly, and this process repeats until convergence.
To support multi-landmark learning, we concatenate orthogonal 2D views (2.5D) and pass them through a shared CNN that regresses the displacement.
Global spatial relationships among landmarks are modeled using a PCA-based shape space. 
Let $\mathbf{X}\in\mathbb{R}^{3n_\ell}$ denote the concatenation of the 3D coordinates of all $n_\ell$=16 landmarks, $\bar{\mathbf{X}}$ the mean landmark configuration computed from the training set, and $\mathbf{W}\in\mathbb{R}^{3n_\ell\times n_b}$ the matrix of principal components. 
Each landmark configuration is expressed as
\begin{equation}
\mathbf{X} = \bar{\mathbf{X}} + \mathbf{W}\mathbf{b},
\label{eq:pca_shape}
\end{equation}
where $\mathbf{b}\in\mathbb{R}^{n_b}$ represents the low-dimensional shape parameters. 
We set $n_b = 36$, retaining approximately $99.5\%$ of the total variance.
During inference, PIN refines the shape parameters $\mathbf{b}_t$ iteratively rather than directly regressing absolute landmark coordinates, starting from the mean shape initialization $\mathbf{b}_0=\mathbf{0}$. 
At iteration $t$, the CNN predicts a displacement vector $\mathbf{d}_t$ and an associated confidence score $\mathbf{P}_t^{\max}$, which are used to update the shape parameters as
\begin{equation}
\mathbf{b}_{t+1} = \mathbf{b}_t + \mathbf{P}_t^{\max}\odot\mathbf{d}_t,
\label{eq:pin_update}
\end{equation}
followed by reconstruction of the landmark coordinates $\mathbf{X}_{t+1}$ using Eq.~\eqref{eq:pca_shape}~\cite{Li2018PIN}. 
The PIN models were trained for 100{,}000 iterations using Adam (learning rate $10^{-3}$) with a joint loss combining Euclidean regression and cross-entropy classification terms.

\emph{Local models.}
Coarse landmark estimates from the global PIN are refined using local models. To simulate realistic inference conditions, local training patches are generated by perturbing the landmark locations with Gaussian noise.
For a given landmark $i$, let $\mathbf{p}_i$ denote its ground-truth position and let $r_i$ be the maximum radial localization error observed when applying the global PIN on the validation set.
We sample a displacement $\boldsymbol{\delta}_i \sim \mathcal{N}(\mathbf{0}, \sigma_i^2 \mathbf{I})$ and center the training patch at
$\mathbf{c}_i = \mathbf{p}_i + \boldsymbol{\delta}_i$.
The standard deviation $\sigma_i$ is chosen such that $95\%$ of the sampled displacements satisfy $\|\boldsymbol{\delta}_i\|_2 \le r_i$.
For each landmark, we train a landmark-specific 3D fully convolutional neural network (FCNN) following the architecture proposed in~\cite{Noothout2020}.
Each network takes as input a $16^3$ patch centered at $\mathbf{c}_i$ and predicts (i) a 3D displacement vector from the patch center to the true landmark location and (ii) a landmark presence score.
All local FCNNs are trained using Adam (learning rate $10^{-3}$) with a joint loss combining mean absolute error on log-transformed displacement vectors and binary cross-entropy for landmark presence.

\emph{Inference.}
The global PIN provides coarse landmark coordinates. For each landmark, a $16^3$ patch is extracted around its predicted position, processed by its corresponding FCNN, and the predicted displacement is added to the patch center to obtain the final landmark coordinate.

\textbf{Step 2 -- 3D semantic segmentation.}
To generate the 12-label coarse subcortical segmentations, we used UNesT~\cite{Yu2023}, a hierarchical 3D transformer that combines nested-window self-attention with U-Net-style skip connections for efficient volumetric segmentation. Additionally, as a baseline, we trained a UNesT model to directly segment the original 26 anatomical labels without any information from landmarks. Both the 12-label (used for refinement) and 26-label (baseline) models were trained for 300 epochs using AdamW (initial learning rate $10^{-4}$, cosine decay schedule) and a compound Dice+cross-entropy loss.

\textbf{Step 3 -- Landmark-guided post-processing.}
As described in Sec.~\ref{sec:hoacoding}, the coarse UNesT output (12 labels) can be deterministically refined into 26 anatomically faithful labels using landmark-driven rules. We follow this procedure as the final step of our landmark-driven segmentation pipeline.

\subsection{Performance evaluation}
Landmark localization accuracy is reported as mean Euclidean error (mm). Segmentation quality is assessed through (i) global overlap (Dice) and (ii) boundary accuracy near protocol-defined planes.

\textbf{Protocol-Aligned Surface Distance (PASD).}
Conventional surface metrics (e.g., average symmetric surface distance (ASSD)~\cite{Heimann2009}) evaluate distances over entire object boundaries and do not focus on the specific separation planes defined by the anatomical protocol. We therefore compute a \emph{Protocol-Aligned Surface Distance (PASD)}, a localized and one-way variant of ASSD that measures the displacement of a protocol-defined surface toward the predicted segmentation.
For each landmark-driven boundary of a neuroanatomical region, the ground-truth surface $\mathcal{S}_{\mathrm{GT}}^{(p)}$ is extracted from the protocol-defined plane (typically coronal). From the prediction, we collect all voxels $\mathcal{V}_{\mathrm{Pred}}^{(s)}$ lying on the corresponding anatomical side. PASD is then defined as
\begin{equation}
\mathrm{PASD}=
\frac{1}{\left|\mathcal{S}_{\mathrm{GT}}^{(p)}\right|}
\sum_{v_i\in\mathcal{S}_{\mathrm{GT}}^{(p)}}
\min_{u_j\in\mathcal{V}_{\mathrm{Pred}}^{(s)}}
\left\|v_i-u_j\right\|_2 .
\label{eq:pasd}
\end{equation}
This asymmetric formulation quantifies how closely the predicted boundary aligns with the anatomically defined plane, rather than global shape similarity.

\textbf{2D separation regularity.}
For slices intersecting landmark-defined planes, we further evaluate the 2D behavior of the predicted separation lines. Let $y_i^{\mathrm{pred}}$ and $y_i^{\mathrm{gt}}$ denote the predicted and reference line positions at point $i$, respectively, and let $N$ be the number of evaluated points within the slice. We compute the mean absolute error (MAE) to assess line-position accuracy, and the within-slice standard deviation $\sigma_y$ to assess the regularity of the predicted line. Here, $\bar{y}_{\mathrm{pred}}$ denotes the mean predicted line position within the considered slice:
\begin{align}
\mathrm{MAE} &= 
\frac{1}{N}\sum_{i=1}^{N}
\left|y_i^{\mathrm{pred}}-y_i^{\mathrm{gt}}\right|,
\label{eq:mae}\\
\sigma_y &=
\sqrt{
\frac{1}{N}\sum_{i=1}^{N}
\left(y_i^{\mathrm{pred}}-\bar{y}_{\mathrm{pred}}\right)^2
}.
\label{eq:sigma}
\end{align}
Thus, MAE measures the accuracy of the separation line location, whereas $\sigma_y$ measures the within-slice variability of the predicted line. Lower $\sigma_y$ values indicate straighter and more regular protocol-defined separations.

\section{Results}
\label{sec:results}

All evaluations were performed on the \emph{same 10} test subjects to enable direct comparisons across models.

\begin{figure*}[t]
    \centering
    \includegraphics[width=0.9\textwidth]{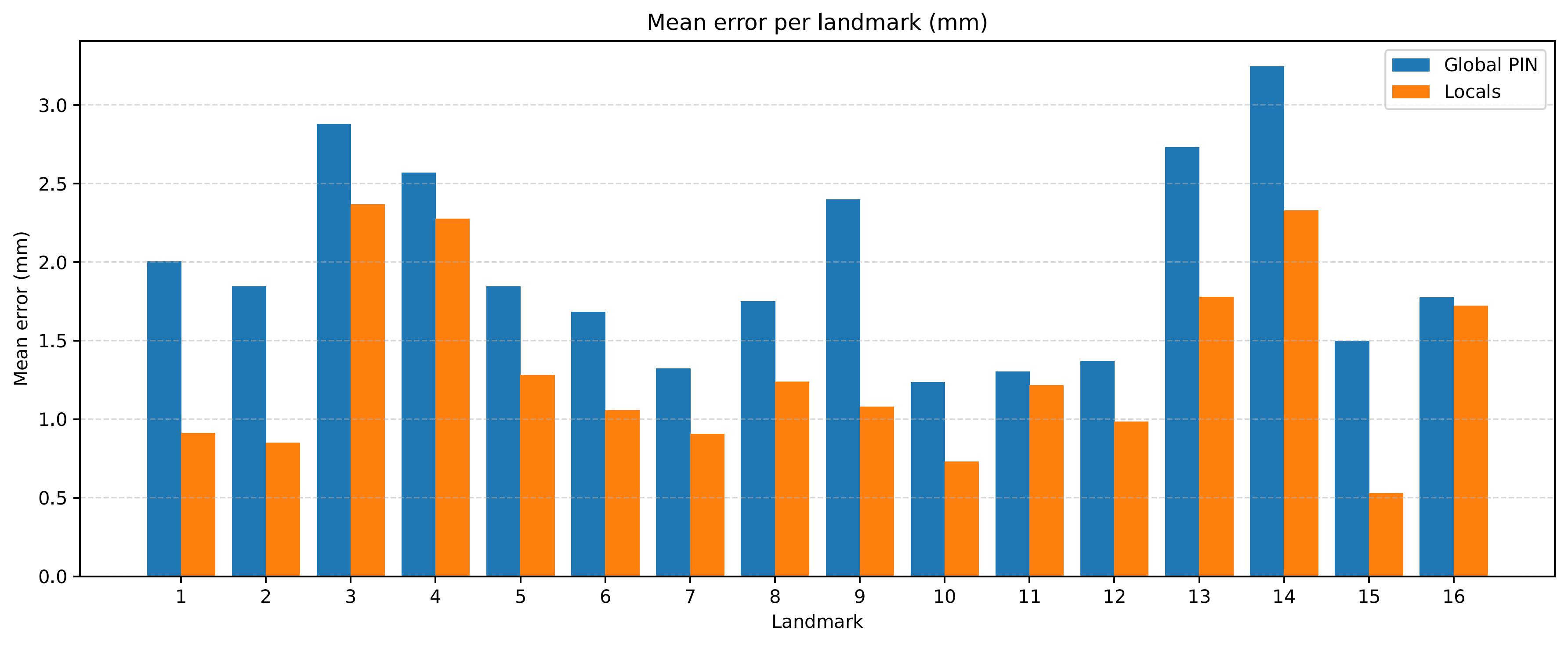}
    \caption{Landmark localization error for the global PIN and local refinement models.}
    \label{fig:landmark_error_hist}
\end{figure*}

\textbf{Landmark detection results.}
\label{para:lanRes}
Figure~\ref{fig:landmark_error_hist} reports the mean localization error for each anatomical landmark, comparing the \emph{global PIN} model with the corresponding \emph{local refinement networks}. The global PIN acts as a coarse localization stage, jointly predicting all landmarks and achieving an overall mean error of \textbf{1.97\,mm}. The local models, trained independently for each landmark, further reduce the mean error to \textbf{1.33\,mm}, corresponding to an improvement of approximately \textbf{32\%}.

Notably, localization accuracy varies across landmarks. Highly distinctive anatomical landmarks, such as the AC and PC (\#10, \#15), are localized with higher precision (approximately $0.6$\,mm) due to their strong image contrast. In contrast, landmarks defining boundaries between anatomically similar regions—such as the continuity of the atrium used to separate the LV and IH (\#13, \#14, $\sim$3\,mm), and the NAcc--Putamen interface (\#3, \#4, $\sim$2.6\,mm)—exhibit higher localization errors. These regions present limited contrast and are primarily defined by protocol-driven criteria, making them inherently more challenging for CNN-based models to localize accurately. 

\begin{figure*}[t]
    \centering
    \includegraphics[width=0.9\textwidth]{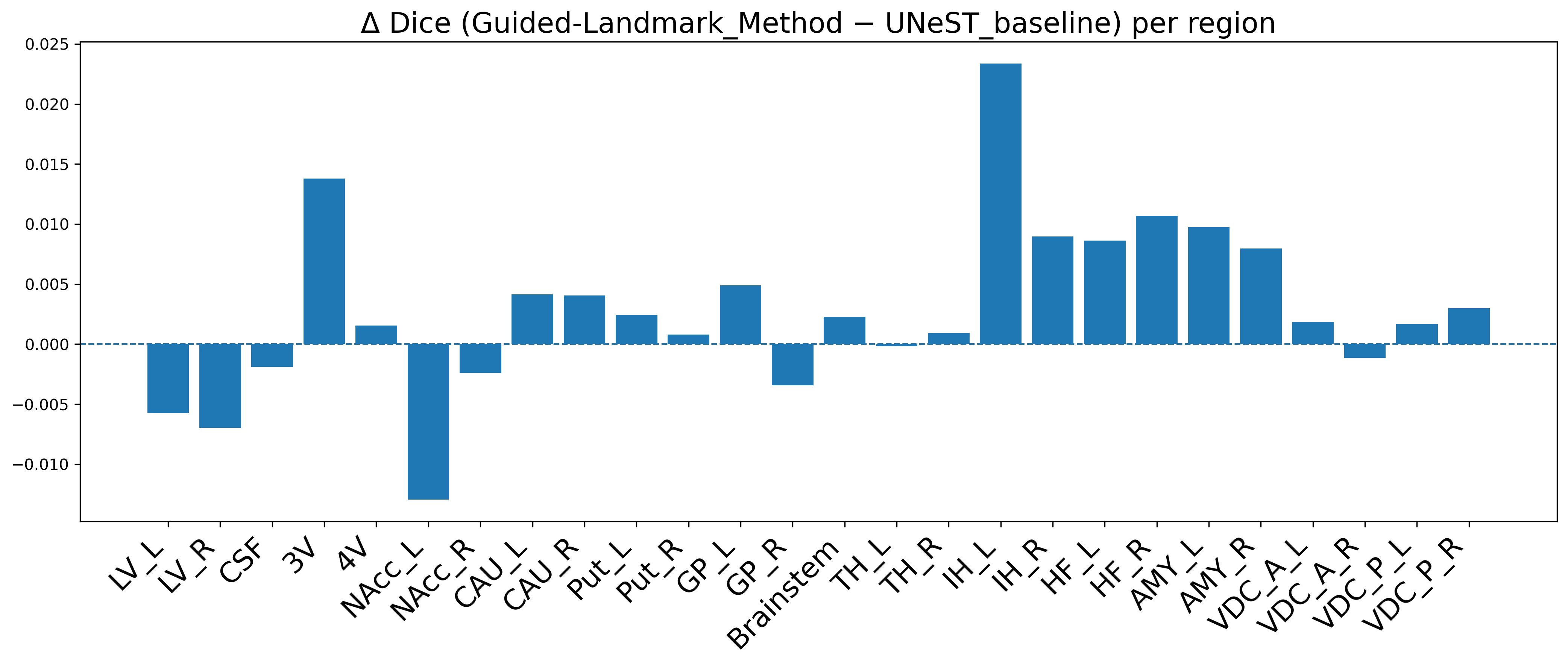}
    \caption{Differences in Dice between landmark-guided and baseline UNesT for each structure.}
    \label{fig:figdice}
\end{figure*}

\textbf{Segmentation results.}
Landmark-guided post-processing yields a modest improvement in global overlap, with mean Dice increasing from $0.8961$ to \textbf{0.8992}. As shown in Figure~\ref{fig:figdice}, Dice differences are generally small ($<0.03$), suggesting that the benefits of landmark-driven segmentation are better captured near protocol-defined boundaries rather than by global volumetric overlap alone.

Some regions exhibit slight decreases in Dice compared to the baseline. These regressions are mainly associated with structures whose separation depends on landmarks that are inherently more difficult to localize, as discussed in the landmark detection results, such as those used to separate the LV from the IH and the NAcc--Putamen interface. As a result, errors in landmark detection may propagate to the refinement stage, leading to local degradations in Dice and highlighting the sensitivity of the refinement step to landmark localization accuracy.

Table~\ref{tab:surface_metrics_condensed} shows that landmark-guided refinement reduces boundary errors in most cases, as measured by the proposed PASD metric, with the best value for each left/right comparison highlighted in bold. For example, on the right inferior horn, PASD decreases from 0.636 to \textbf{0.356}. However, slight increases are observed for a few regions, which can be attributed to the propagation of localization errors from more challenging landmarks.

\begin{table}[htbp]
\caption{PASD (lower is better) for relevant protocol-defined boundaries. A: UNesT baseline; B: landmark-guided.}
\label{tab:surface_metrics_condensed}
\begin{center}
\setlength{\tabcolsep}{4pt}
\renewcommand{\arraystretch}{1.1}
\begin{tabular}{|l|c|c|c|c|}
\hline
\textbf{Region (Surface)} & \multicolumn{2}{c|}{\textbf{Left}} & \multicolumn{2}{c|}{\textbf{Right}} \\
\cline{2-5}
 & \textbf{A} & \textbf{B} & \textbf{A} & \textbf{B} \\
\hline
IH (posterior)        & 0.787 & \textbf{0.524} & 0.636 & \textbf{0.356} \\
\hline
NAcc (lateral)        & \textbf{0.597} & 0.651 & \textbf{0.364} & 0.370 \\
\hline
NAcc (posterior)      & 0.427 & \textbf{0.388} & \textbf{0.398} & 0.449 \\
\hline
Put (anterior)        & 0.491 & \textbf{0.470} & \textbf{0.302} & 0.480 \\
\hline
Put (lateral)         & 0.451 & \textbf{0.424} & 0.459 & \textbf{0.441} \\
\hline
3V (anterior)         & 0.601 & \textbf{0.514} & -- & -- \\
\hline
VDC\_A (posterior)    & 0.650 & \textbf{0.465} & 0.384 & \textbf{0.352} \\
\hline
VDC\_P (anterior)     & 0.444 & \textbf{0.413} & \textbf{0.331} & 0.363 \\
\hline
\end{tabular}
\end{center}
\end{table}
Table~\ref{tab:2d_metrics_compact} complements the PASD analysis by reporting the 2D line-position error (MAE) and within-slice standard deviation ($\sigma_y$), with the best value for each left/right comparison highlighted in bold. The landmark-guided method yields mixed and generally marginal changes in MAE, indicating comparable boundary positioning accuracy to the baseline. In contrast, it consistently reduces $\sigma_y$ to near-zero values, showing that the predicted separation lines are straighter, more stable within each slice, and more compliant with the landmark-defined protocol.

\begin{table}[htbp]
\caption{2D line-position and regularity metrics. A: UNesT baseline; B: landmark-guided. Lower values are better.}
\label{tab:2d_metrics_compact}
\begin{center}
\setlength{\tabcolsep}{4pt}
\renewcommand{\arraystretch}{1.1}
\begin{tabular}{|l|c|c|c|c|}
\hline
\textbf{Region (Surface)} & \multicolumn{2}{c|}{\textbf{Left}} & \multicolumn{2}{c|}{\textbf{Right}} \\
\cline{2-5}
 & \textbf{A} & \textbf{B} & \textbf{A} & \textbf{B} \\
\hline
\multicolumn{5}{|l|}{\textbf{MAE}} \\
\hline
IH (posterior)        & \textbf{0.706} & 0.903 & 0.889 & \textbf{0.471} \\
\hline
VDC\_P (anterior)     & 0.619 & \textbf{0.614} & \textbf{0.480} & 0.488 \\
\hline
NAcc (lateral)        & \textbf{0.826} & 0.913 & \textbf{0.867} & 0.945 \\
\hline
\multicolumn{5}{|l|}{$\boldsymbol{\sigma_y}$} \\
\hline
IH (posterior)        & 0.110 & \textbf{0.000} & 0.167 & \textbf{0.049} \\
\hline
VDC\_P (anterior)     & 0.153 & \textbf{0.000} & 0.242 & \textbf{0.000} \\
\hline
NAcc (lateral)        & 0.118 & \textbf{0.000} & 0.095 & \textbf{0.000} \\
\hline
\end{tabular}
\end{center}
\end{table}

\textbf{Ablation study.}
To further evaluate the impact of landmark localization and post-processing, we compared three landmark configurations: ground-truth (GT) landmarks, locally refined landmarks, and global PIN predictions without local refinement. We also evaluated whether the proposed landmark-guided rules can reconstruct the original 26-label annotation from fused labels independently of segmentation errors.

As shown in Table~\ref{tab:ablation_landmark_sources}, applying the proposed rules to GT fused labels yields near-perfect reconstruction when GT landmarks are used, confirming that the deterministic landmark-guided rules accurately reproduce the HOA protocol.

When applied to the UNesT coarse segmentation, GT landmarks produce the largest improvements in both Dice and PASD, while locally refined landmarks consistently outperform global PIN predictions alone. The least accurate results are observed for the UNesT baseline, where no landmark-guided refinement is applied, and for the coarse segmentation refined using only global landmark predictions.
Overall, these results demonstrate that accurate landmark localization is critical for enforcing protocol-aligned boundaries and further highlight the importance of the local refinement stage in the proposed framework.
\begin{table}[htbp]
\caption{Landmark-guided segmentation using different configurations. Each cell reports Dice ($\uparrow$) / PASD ($\downarrow$) after applying the post-processing. $^{*}$ indicates a statistically significant improvement of the corresponding metric compared with the UNesT baseline after paired Wilcoxon signed-rank tests with FDR correction~\cite{wilcoxon1945,benjamini1995}.}
\label{tab:ablation_landmark_sources}
\begin{center}
\scriptsize
\setlength{\tabcolsep}{2.2pt}
\renewcommand{\arraystretch}{1.12}
\begin{tabular}{|p{0.30\columnwidth}|c|c|c|}
\hline
\textbf{Segmentation input} &
\multicolumn{3}{c|}{\textbf{Input landmarks}} \\
\cline{2-4}
&
\textbf{GT} &
\textbf{Local models} &
\textbf{Global PIN} \\
\hline

GT fused labels &
0.9965$^{*}$ / 0.0720$^{*}$ &
0.9926$^{*}$ / 0.2789$^{*}$ &
0.9891$^{*}$ / 0.4231$^{*}$ \\
\hline

UNesT coarse seg. &
0.9026$^{*}$ / 0.4558$^{*}$ &
0.8992$^{*}$ / 0.6339 &
0.8957 / 0.7735 \\
\hline

UNesT baseline &
\multicolumn{3}{c|}{0.8961 / 0.6975} \\
\hline

\end{tabular}
\end{center}
\end{table}
\textbf{Qualitative evaluation.}
Five randomly selected cases were qualitatively assessed by R.~J.~Rushmore, co-author of the HOA protocol. He reported that the landmark-guided method adheres more closely to the protocol than the baseline model.
In particular, the landmark-defined planes correctly
enforce anterior/posterior truncation and left/right separation
of structures, whereas the baseline often violates these boundaries.

\begin{figure*}[t]
    \centering
    \includegraphics[width=0.85\textwidth]{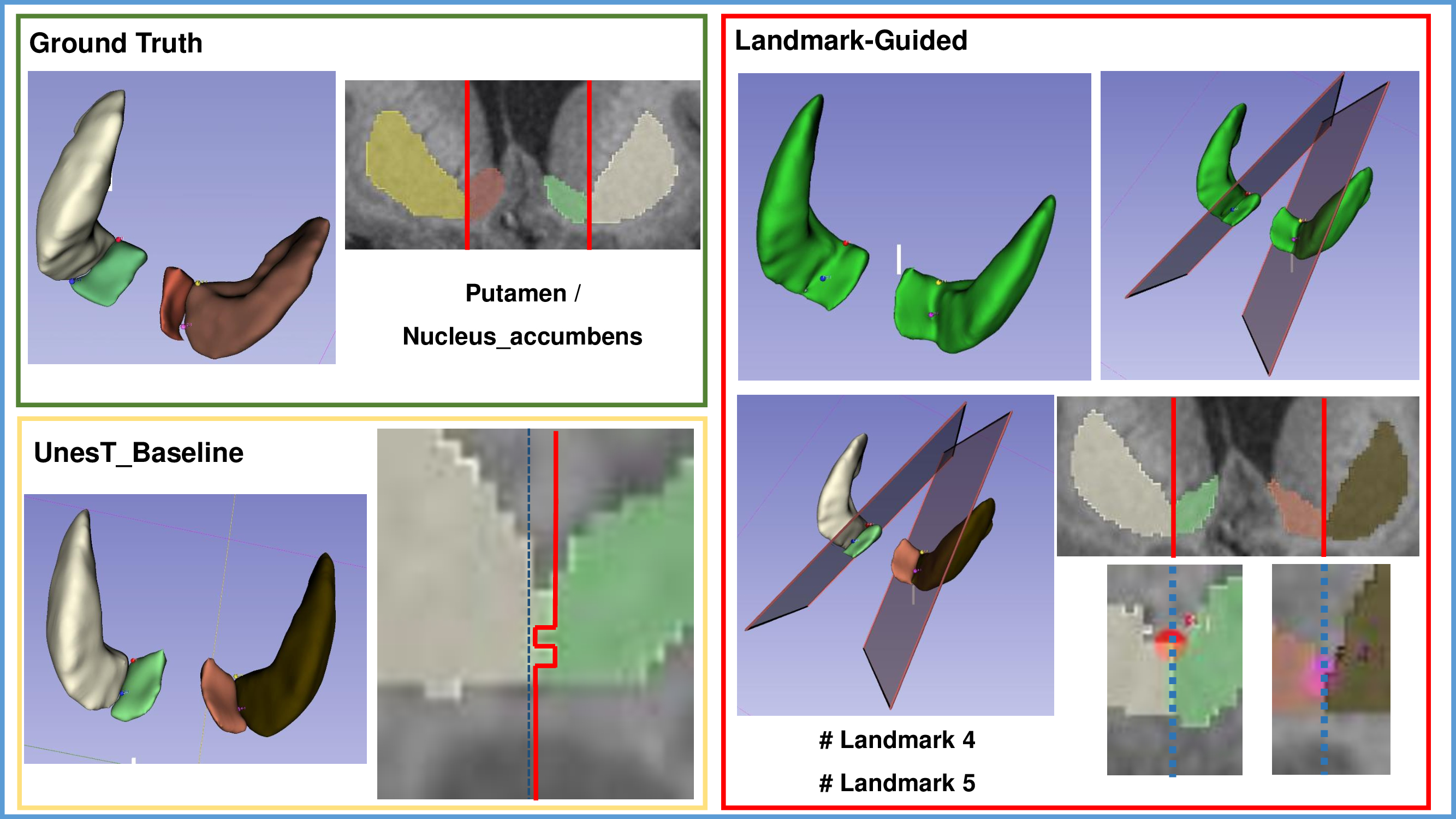} \\[2pt]
    \includegraphics[width=0.85\textwidth]{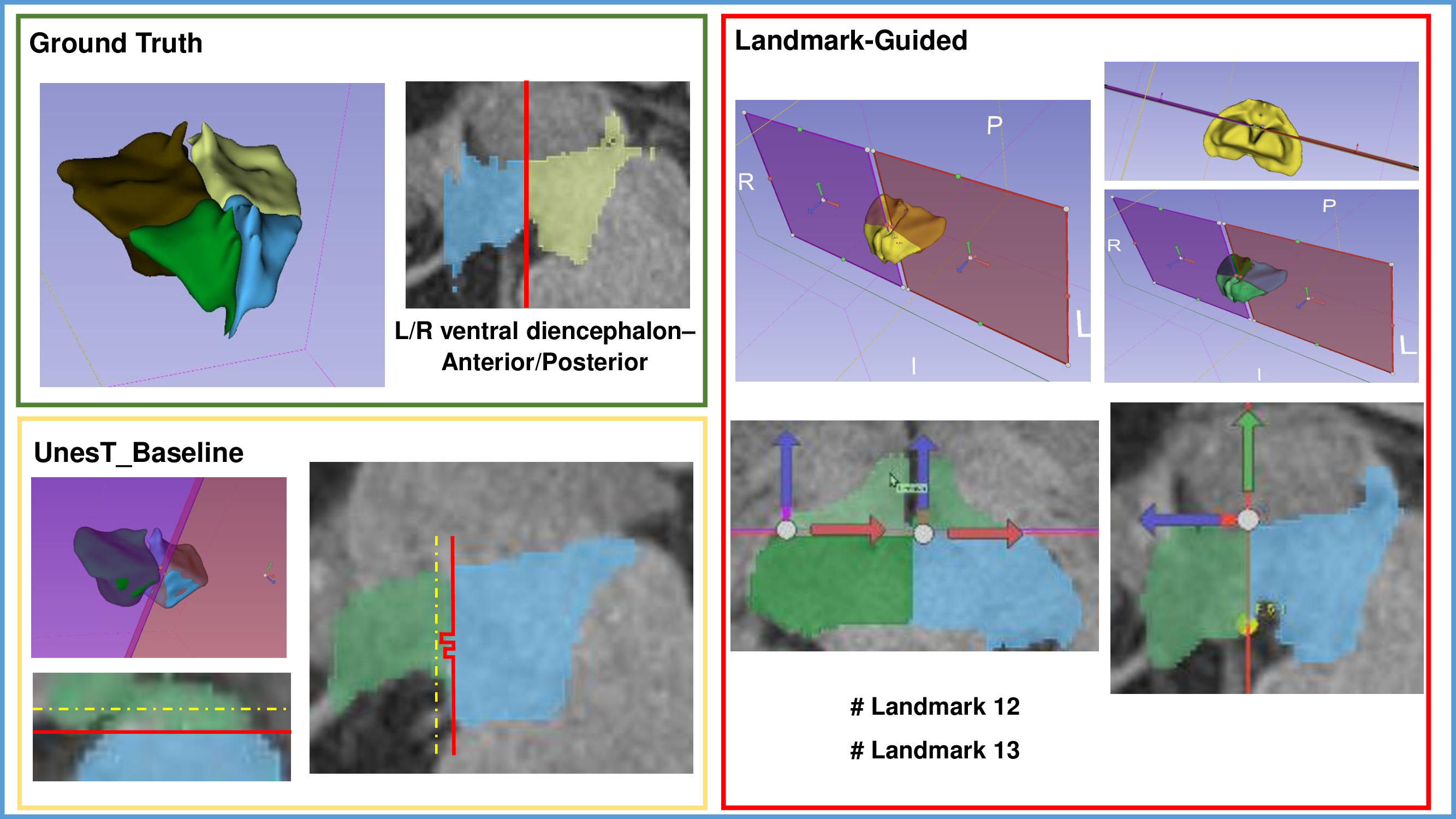}
    \caption{\textbf{Top:} Landmark-guided separation of the putamen and nucleus accumbens using contact landmarks (\#3--\#6). 
    \textbf{Bottom:} Landmark-defined coronal plane (based on mammillary bodies, \#11--\#12) splitting anterior and posterior ventral diencephalon. }
    \label{fig:expert_quals}
\end{figure*}

Figure~\ref{fig:expert_quals} illustrates two typical cases where the landmark-guided method better follows protocol-defined boundaries than
the UNesT baseline. In the first row, it correctly separates
the putamen from the nucleus accumbens using contact landmarks,
whereas the baseline produces a fused and irregular
interface. In the second row, the mammillary body landmarks
enable proper anterior–posterior separation of the ventral diencephalon,
which the baseline fails to enforce. These corrections
are local but clinically meaningful, as they improve
structural delineation without altering global metrics.

\section{Conclusion and future directions}
In this work, we presented a landmark-guided framework for automatic segmentation of subcortical structures. 
By combining landmark detection with a coarse UNesT segmentation and simple post-processing rules, the method enforces boundary placements that are consistent with neuroanatomical protocols. 
On the HOA dataset, gains in Dice score were modest, but boundary accuracy near landmarks improved markedly.
Importantly, the expert neuroanatomist consistently preferred the landmark-guided framework over the pure UNesT segmentation as it mimicked the manual protocol more closely.

This study is limited to the HOA subcortical dataset derived from the Human Connectome Project, which consists primarily of healthy young adults. This choice is motivated by the fact that the HOA dataset is the only available resource with manual annotations explicitly following the HOA neuroanatomical protocol, which we aim to automate. While this restricts direct evaluation to a specific population, the proposed framework is inherently protocol-driven and can be extended to other datasets and anatomical protocols, provided that the corresponding landmarks and rules are defined.

Future work could extend these principles to other segmentation tasks, and adapt them for semi-automated segmentation protocols.
Additionally, recent work on landmark localization using foundation models such as MedLAM~\cite{Lei2025MedLSAM} could be explored to increase generalizability. However, the landmarks considered here are expert-defined, protocol-driven neuroanatomical reference points that delineate boundaries between adjacent subcortical regions and often lack distinctive local image descriptors which can be challenging for these foundational models.

Overall, our work identified a gap between automated and manual neuroanatomical annotation protocols, which is not currently addressed by direct semantic segmentation, and emphasizes the need for additional information via landmarks to reproduce manual protocols most accurately. 

\section*{Ethics and Acknowledgements}
\label{sec:ethics}
This research used retrospective human subject data from the Human Connectome Project. 
The experimental procedures involving human subjects described in this paper were approved by the Ethics Committee of ÉTS. 
This research is supported by the Canada Research Chairs and NSERC Discovery programs.

\end{document}